%% file: acl_latex.tex
\documentclass[11pt]{article}

\usepackage[preprint]{acl}

\usepackage{times}
\usepackage{latexsym}

\usepackage[T1]{fontenc}

\usepackage[utf8]{inputenc}

\usepackage{microtype}

\usepackage{inconsolata}

\usepackage{graphicx}
\usepackage{booktabs}

\usepackage{subcaption}
\usepackage{wrapfig}
\usepackage[most]{tcolorbox}
\tcbuselibrary{breakable}
\usepackage{xcolor} 
\usepackage{caption}
\usepackage{amsfonts}       
\usepackage{nicefrac}       
\usepackage{microtype}      
\def\method{ORACLE}

\title{\method{}: Anticipating Scams from Partial Trajectories in Streaming App Usage}

\author{
\textbf{
Wenbo Gao$^{1}$,
Songbai Tan$^{2}$,
Zhongan Wang$^{3}$,
Fei Shen$^{4}$,
Gang Xu$^{5}$
} \\
\vspace{-0.2cm}
\textbf{
Huiping Zhuang$^{6}$,
Yunyun Yang$^{1}$\thanks{Corresponding author: \texttt{yangyunyun@hit.edu.cn}},
Ming Li$^{5}$\thanks{Corresponding author: \texttt{liming@gml.ac.cn}},
Xiaofeng Zhu$^{7}$
} \\
\\
$^{1}$Harbin Institute of Technology, Shenzhen 
$^{2}$Shenzhen University \\
$^{3}$Zhejiang University
$^{4}$National University of Singapore 
$^{5}$Guangming Laboratory \\
$^{6}$South China University of Technology 
$^{7}$Hainan University
}

\input{math_commands.tex}

\begin{document}
\maketitle
\begin{abstract}
\input{sec/abstract}
\end{abstract}

\input{sec/introduction}
\input{sec/dataset}
\input{sec/method}
\input{sec/exp}
\input{sec/conclusion}

\section*{Limitations}
Due to the difficulty of collecting large-scale, fully observed victim-side app-usage histories, our evaluation relies on a benchmark curated from real-world normal app usage data and real scam cases.
Although the benchmark constructs long-horizon trajectories with interleaved normal and scam behaviors across diverse scam types, it may not fully capture the complexity and distribution shifts of real-world victim interactions.
Future access to privacy-preserving, naturally collected victim-side trajectories would enable more faithful evaluation and stronger validation in real deployment settings.

\newpage


\bibliography{anthology}
\newpage
\appendix

\input{sec/appendix}
\label{sec:appendix}

\end{document}

%% file: math_commands.tex
\usepackage{amsmath,amsfonts,bm}

\def\eqref#1{equation~\ref{#1}}

\def\1{\bm{1}}

\def\rvh{{\mathbf{h}}}

\def\rvs{{\mathbf{s}}}

\DeclareMathAlphabet{\mathsfit}{\encodingdefault}{\sfdefault}{m}{sl}
\SetMathAlphabet{\mathsfit}{bold}{\encodingdefault}{\sfdefault}{bx}{n}

\def\gE{{\mathcal{E}}}

\def\gM{{\mathcal{M}}}

\def\gS{{\mathcal{S}}}

\def\gW{{\mathcal{W}}}



%% file: sec/abstract.tex
Smartphone scams are increasingly prevalent and typically manifest as multi-stage, cross-application processes with gradually emerging intent. Effective intervention thus requires anticipating scams before the intent becomes explicit. This is inherently challenging, as decisions must rely on partial trajectories with temporally distributed evidence.
In this paper, we propose \textbf{\method{}} Online Reasoning for Anticipating Cross-temporal Latent thrEats, the first agentic framework for early scam anticipation from \textit{streaming app-usage} trajectories.
To support this setting, we curate a real-world long-horizon benchmark of streaming app-usage trajectories, covering 12 scam types, spanning extended periods (15 days on average), involving diverse applications (95 apps), and interleaving normal and scam behaviors. 
%
To address fragmented evidence, we introduce a self-evolving context manager that adaptively consolidates entity-centric interactions over time, enabling more effective reconstruction of cross-temporal evidence from partial observations.
To enhance sensitivity to latent early-stage signals, we propose an on-policy skill-guided distillation scheme in which a teacher model, conditioned on summarized anti-scam reflections and clues by skills, supervises a student model without access to such reflections. This scheme thereby distills evidence-informed knowledge and improves recognition of emerging fraud patterns from partial trajectories.
Experiments show that \method{} consistently improves early scam anticipation, yielding timely warnings while reducing false alerts in realistic streaming scenarios.

%% file: sec/introduction.tex
\section{Introduction}
\begin{figure*}
    \centering
    \includegraphics[width=0.95\linewidth]{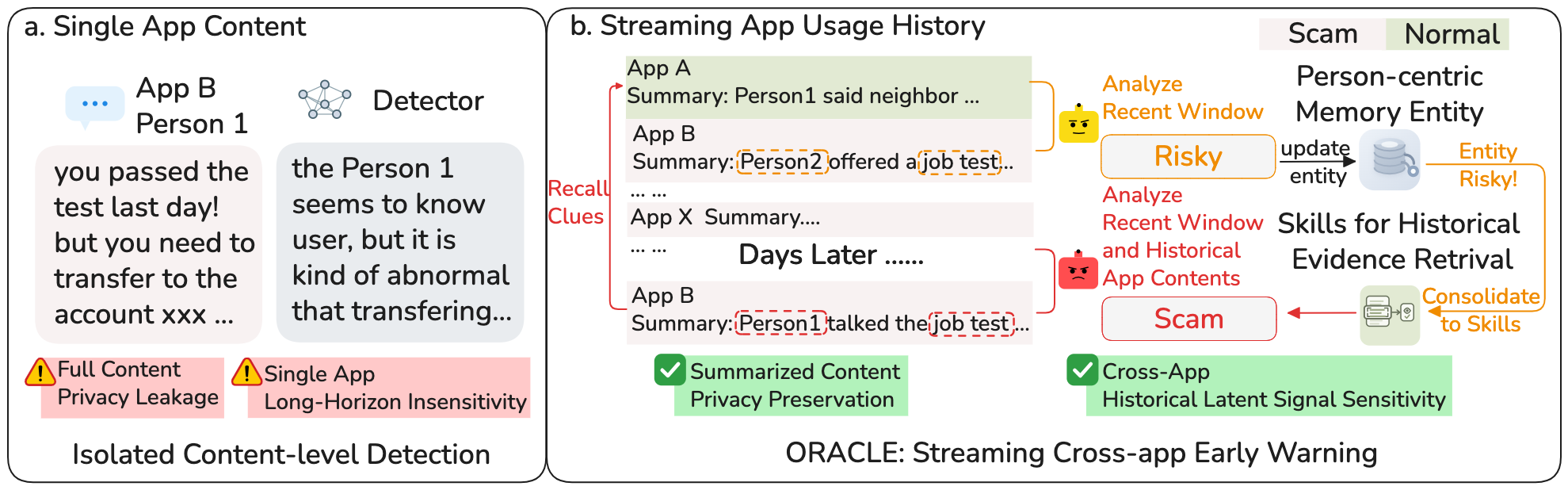}
\caption{
\textbf{Comparison between isolated content-level detection and streaming cross-app anticipation.}
(a) Existing methods analyze single app sessions independently, which is insufficient for long-term scam processes where historical evidence is distributed across apps and time.
(b) \method{} analyzes app interactions in the recent window and uses a memory-skill context manager to retrieve entity-related historical evidence from previous trajectories, enabling cross-temporal reasoning for earlier scam-risk anticipation.
}
\vspace{-7mm}
    \label{fig:teaser}
\end{figure*}

Smartphones concentrate communication, social interaction, and financial transactions within a single device, making them a natural attack surface for remote scams~\cite{tan2024scamgpt,ahvanooey2017survey} and posing significant risks to daily life and financial security. The investigation reveals that telecom scams have surged, causing substantial financial losses and increasing threats to personal safety \cite{agarwal2025overview,tan2025anticipate,shen2025warned}.

Therefore, scam detection has attracted increasing attention from the research community~\cite{tan2024scamgpt,basta2025bot,jaipuria2025case}. Most prior work focuses on analyzing communication content within \textit{isolated applications}, such as phone calls or message threads \cite{fraudr1,ma2025teleantifraud,agarwal2025overview,tan2024scamgpt,safeqaq}. 
For example, ScamGPT-J~\cite{tan2024scamgpt} addresses instant-messaging scam detection with a fine-tuned large language model that simulates scammer responses in real time, helping users identify suspicious interactions through analogy-based reasoning. TeleAntiFraud~\cite{ma2025teleantifraud} proposes a slow-thinking multimodal framework for voice-based telecom fraud analysis, and establishes a fine-tuned Qwen2-Audio~\cite{chu2024qwen2} baseline for fraud-type identification from telephone audio. 


However, real-world scams rarely appear as isolated malicious events. Prior sociological studies indicate that scams typically unfold as a \textit{temporal multi-stage multi-application process}, including calls, messages, social platforms, and financial services~\cite{li2026linguistic,DULISSE2026100211}. Fraudulent intent is gradually revealed through a sequence of weak behavioral cues, while the window for effective intervention remains short. For example, scammers may initiate contact through one channel, establish trust through another, and eventually induce financial actions through yet another \cite{li2026linguistic}. As a result, the underlying risk signal is temporally distributed rather than localized, making early anticipation from streaming app-usage trajectories significantly more challenging than one-shot classification.
On the other hand, isolated application-based anti-scam relies heavily on detailed content analysis \cite{tan2024scamgpt,ma2025teleantifraud,chu2024qwen2}, which raises significant privacy concerns, limiting its practicality in real-world deployment.

In this work, we propose early scam intervention from streaming app usage. This setting can be naturally formulated as a problem of reasoning under incomplete observations, and poses two key challenges. First, \textbf{fragmented context}: scam-risk anticipation operates in a streaming manner over continuously evolving app-usage trajectories, where intent clues are fragmented and distributed across long temporal horizons \cite{li2026linguistic,DULISSE2026100211}. Interpreting current behavior therefore requires cross-temporal reasoning over historical interactions beyond the visible window. Second, \textbf{latent early signals}: in the early stages of a scam, malicious intent is often deliberately obscured and embedded within normal user behavior, making it difficult to distinguish emerging fraud patterns from benign activity based on partial observations \cite{li2026linguistic}.

To address these challenges, we propose \textbf{\method{}}, an agentic framework of Online Reasoning for Anticipating Cross-temporal Latent thrEats, enabling early scam anticipation from streaming app-usage trajectories.
To enable this setting, we curate a long-horizon benchmark of streaming app-usage trajectories, constructed from \textit{real telecom scam cases and criminal records}. The dataset features extended temporal spans (15 days on average), diverse application interactions (95 apps grouped into 12 categories), and interleaved normal and scam behaviors. It consists of 57,662 short traces aggregated into 3,061 long trajectories, each averaging 96 app events and covering 12 scam types, presenting a challenging testbed for early scam anticipation under realistic conditions.
%

Building upon this benchmark, we design \method{} from both system and learning perspectives. 
At the system level, we introduce a \textit{self-evolving context manager} that progressively refines entity-centric memory from streaming interactions and decision feedback, enabling increasingly accurate reconstruction of temporally dispersed evidence beyond the local observation window.
To enhance sensitivity to latent early-stage signals, we propose an \textit{On-Policy Skill-guided Distillation (OPSD)} that leverages generated anti-scam reflections and clues as privileged information. Specifically, a teacher model conditioned on these reflections provides supervision, while the student model operates without access to them. By aligning the student’s predictions with the teacher, the model effectively internalizes reflection-informed knowledge, improving its ability to recognize emerging fraud patterns from partial trajectories.

Our main contributions are as follows:
\vspace{-3mm}

\begin{itemize}
\item We propose a self-evolving agent that performs cross-temporal reasoning to bridge scattered fraudulent clues. By selectively retrieving historical interactions across various applications, this mechanism effectively addresses the challenge of fragmented evidence in long-horizon scam scenarios.
\vspace{-3mm}
\item We introduce an On-Policy Skill-guided Distillation (OPSD) scheme to sharpen the model's sensitivity toward subtle, early-stage scam patterns. By encoding expert-derived anti-scam skills into textual scam lessons and distilling them into model parameters, the agent learns to identify deceptive intent from partial trajectories.
\vspace{-3mm}
\item We curate a long-horizon app-usage benchmark characterized by streaming, cross-app interaction sequences. This dataset provides a realistic testbed for evaluating the proactive warning capabilities of agent systems before scams reach a critical escalation point.
\end{itemize}


%% file: sec/dataset.tex
\section{Streaming Scam Benchmark}
\label{sec:benchmark}

Existing anti-scam benchmarks~\cite{fraudr1,ma2025teleantifraud} primarily evaluate isolated content from a single application, whereas real-world scams often unfold over extended periods through a sequence of weak, cross-app behavioral cues. This mismatch makes existing benchmarks insufficient for studying \emph{early anticipation}: by the time an isolated message, call, or payment becomes explicitly malicious, the opportunity for intervention may already be limited.

To bridge this gap, we construct a long-horizon streaming app-usage benchmark that preserves the temporal evolution of scam behaviors while embedding them within substantial everyday activity. The benchmark combines detailed victim-reported scam records from CCL2023~\cite{sun-etal-2023-ccl23} with real-world app-usage traces from LSApp~\cite{lsapp}. CCL2023 provides rich natural-language descriptions of the complete scam process, including involved applications, temporal progression, anonymized persons, operations, and conversational context, whereas LSApp provides structured everyday app usage with application names, temporal order, and three coarse action states: \texttt{Open}, \texttt{Close}, and \texttt{Interaction}.

We use the LSApp representation as the basis for a unified event schema:
\vspace{-3mm}

\[
(\texttt{order},\ \texttt{app\_category},\ \texttt{action},\ \texttt{description}),
\]

\noindent where exact applications are mapped to 12 functional categories and each event contains a concise operation or interaction description. CCL2023 and auxiliary LOCOMO~\cite{locomo} content are processed through the same summarization pipeline to maintain consistent textual granularity across scam and normal events.

A key characteristic of our benchmark is that scam evidence does not appear as a single isolated malicious block. Instead, each scam process is decomposed according to its original behavioral progression and sparsely interleaved with normal app usage while preserving the relative order of scam events. The resulting trajectories therefore require models to reason over \emph{partial, temporally fragmented, and progressively emerging evidence}, closely matching the target setting of streaming scam anticipation. Full construction procedures, source-normalization details, quality control, and benchmark statistics are provided in Appendix~\ref{app:benchmark}.

\noindent\textbf{Evaluation Metrics.}
Streaming scam anticipation requires more than eventually identifying a scam:
a practical detector should raise warnings from weak latent signals before fraudulent intent becomes explicit, integrate evidence dispersed across time and applications, avoid spurious alerts, and remain reliable under partial observation.
We therefore report four complementary metrics that characterize detection coverage, false alerts, anticipation earliness, and recognition consistency.

\textbf{Hit Rate (HR)} measures whether a scam trajectory is successfully detected at least once during the scam process, while \textbf{False Alert Rate (FAR)} measures the frequency of positive predictions outside the scam process.
Together, they characterize overall detection coverage and false-alert control.

Beyond HR and FAR, we introduce two new temporal metrics tailored to streaming scam anticipation.
\textbf{Earliest Detection Position (EDP)} measures how early the first valid positive prediction occurs within the scam process.
The position of this first alert is normalized by the scam-process length so that trajectories with different temporal extents are comparable, while trajectories that are never detected are assigned the maximum value.
A lower EDP therefore indicates that the model can recognize weak scam signals earlier, before the fraudulent process becomes fully explicit.

\textbf{Partial-observation Alert Rate (PAR)} measures whether early recognition remains consistent as partial scam evidence becomes observable.
For each scam trajectory, we consider windows that end within the scam process and contain at least half of the maximum portion of that process that can be observed within a single window.
PAR is the proportion of these partial-observation candidate windows that are predicted as scam.
Unlike HR, which can be satisfied by a single correct alert, PAR rewards models that continue to recognize the scam across successive incomplete observations.
A higher PAR therefore indicates more stable recognition under partial information.

Together, EDP and PAR capture two distinct properties of early warning:
EDP measures \emph{how early} the first alert is raised, whereas PAR measures \emph{how consistently} the scam is recognized before the complete process is observed.
For evaluation, both \textsc{Risky} and \textsc{Scam} are mapped to the positive scam class, while \textsc{Normal} is treated as negative.
Formal definitions are provided in Appendix~\ref{app:metrics}.

%% file: sec/method.tex
\section{\method{}}
\label{sec:method}
\vspace{-2mm}

\begin{figure*}[t]
    \centering
    \includegraphics[width=0.95\linewidth]{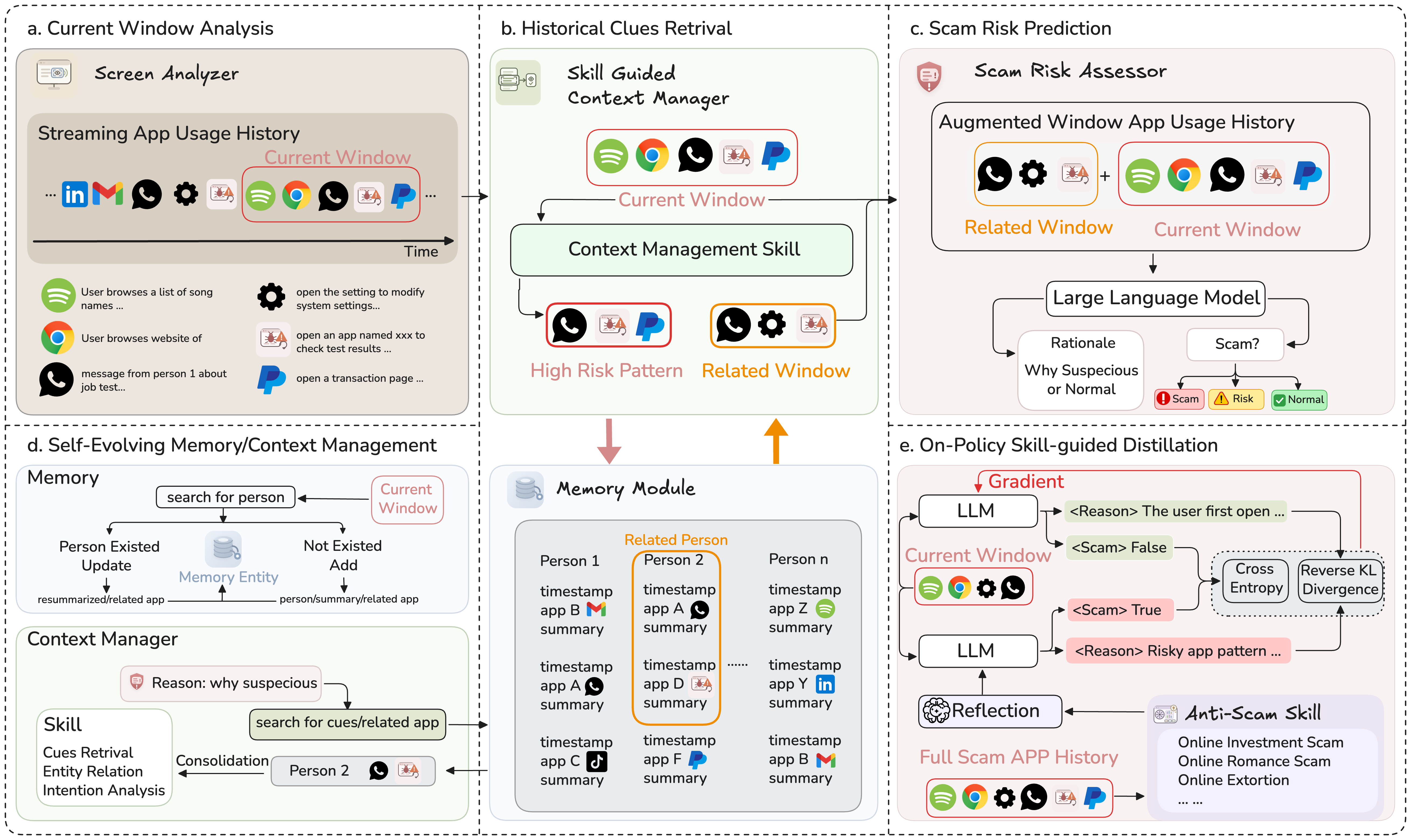}
    \caption{
\textbf{Overview of \method{}.} The pipeline first extracts information from the current window (\textbf{a}), then a skill-guided context manager retrieves related historical interactions to form an augmented window (\textbf{b}), and finally a scam-risk assessor outputs a risk judgment from this augmented context (\textbf{c}). During training, the system undergoes a self-evolving process where memory is updated with incoming events and the skill library is refined from risky/scam predictions (\textbf{d}), as well as on-policy skill-guided distillation where anti-scam reflections are generated by comparing partial and full scam app-usage traces under scam-type-specific skills (\textbf{e}).
    }
    \label{fig:framework}
    \vspace{-4mm}
\end{figure*}

We present \method{}, a streaming agentic framework for early scam anticipation from partial app-usage trajectories.
As shown in Figure~\ref{fig:framework}, ORACLE consists of a \textit{screen analyzer} that extracts entity-centric information, a person-centric \textit{memory store} that maintains historical interactions, a \textit{skill-guided context manager} that retrieves relevant history, and a \textit{scam risk assessor} that produces risk judgments.
During deployment, memory and retrieval skills co-evolve from incoming events and assessor feedback~\cite{jiang2026xskill,xia2026skillrl}, while the assessor parameters remain fixed.
The assessor is trained with \textbf{On-Policy Skill-guided Distillation (OPSD)}, where a stronger teacher with privileged anti-scam reflections supervises a compact student on partial trajectories.

Let $\mathcal{H}=\{\mathbf{h}_1,\ldots,\mathbf{h}_T\}$ denote a streaming app-usage trajectory.
At timestep $t$, the system observes only
$\mathcal{W}_t=\{\mathbf{h}_{t-W+1},\ldots,\mathbf{h}_t\}$.
The policy predicts a binary scam label $y_t\in\{0,1\}$.
For $y_t=1$, we further distinguish \textsc{Risky} and \textsc{Scam} according to prediction confidence, representing low- and high-confidence scam states.
This allows weak early-stage signals to contribute to context evolution before sufficient evidence accumulates for a high-confidence scam judgment.

\subsection{Self-Evolving Context Manager}

Scam evidence is often distributed across applications and may fall outside a fixed observation window~\cite{shen2025warned}.
ORACLE therefore maintains a person-centric memory $\gM_t$ and a skill library $\gS_t$ to retrieve relevant historical evidence beyond the current window.
The person-centric organization is particularly suitable for scams because interactions involving the same contact, account, or service may recur across different applications and become informative only when considered jointly.

A Screen Analyzer $\phi_{\mathrm{scr}}$ extracts entities $\gE_t$ and a concise summary $\rvs_t$ from each incoming event $\rvh_t$, which are stored chronologically in $\gM_t$.
At inference time, entities in $\gW_t$ retrieve associated interactions, while $\gS_t$ provides reusable anti-scam heuristics for ranking the retrieved history.
The augmented observation is
\begin{equation}
\tilde{\gW}_t
=
\textsc{Concat}\!\left(
\textsc{Rank}(\gM_t,\gE(\gW_t),\gS_t),
\gW_t
\right).
\label{eq:augmented-window}
\end{equation}

The context evolves at two timescales.
Memory is updated with every incoming event, whereas high-suspicion rationales and identified cues periodically refine the skill library.
Unlike raw memory accumulation, skill evolution abstracts recurring evidence patterns into reusable retrieval guidance, helping the context manager prioritize historically relevant events rather than simply injecting more history.
This enables ORACLE to progressively improve cross-temporal evidence selection without updating the base model during deployment.
Detailed memory organization and skill evolution are provided in Appendix~\ref{app:context-manager}.

\subsection{On-Policy Skill-guided Distillation}

Early scam windows often contain weak cues whose significance becomes apparent only after later events are observed.
Standard label-based SFT therefore provides limited guidance on why an incomplete early-stage pattern should already be considered suspicious.
We introduce \textbf{On-Policy Skill-guided Distillation (OPSD)} to transfer such future-informed anti-scam knowledge from a stronger teacher to the compact scam assessor.

For a scam-related training window $\tilde{\gW}_t$, the student $\pi_\theta$ observes only the partial augmented trajectory, matching the information available during inference.
A frozen teacher $\pi_{\mathrm T}$ additionally receives a privileged reflection $E_t$, generated by comparing the visible partial trace with the corresponding complete scam trace under a scam-type-specific skill.
The reflection summarizes how currently weak cues may develop into later fraudulent behavior, thereby converting future scam progression into explicit training-time supervision.

Rather than distilling from fixed teacher demonstrations, OPSD performs supervision on token trajectories generated by the current student policy.
The teacher evaluates the same trajectory while additionally conditioning on $E_t$, so the supervision remains aligned with outputs actually visited by the evolving student.
Distillation is performed directly in autoregressive token space through
\begin{equation}
\mathcal{L}_{\mathrm{KL}}
=
D_{\mathrm{KL}}\!\left(
\pi_{\theta}(\cdot\mid\tilde{\gW})
\;\|\;
\pi_{\mathrm T}(\cdot\mid\tilde{\gW},E)
\right),
\label{eq:kl}
\end{equation}
where reverse KL encourages the student to concentrate on high-probability reasoning paths identified by the privileged teacher~\cite{zhao2026self}.

For scam-related windows, we combine distillation with grounded decision supervision:
\begin{equation}
\mathcal{L}
=
\mathcal{L}_{\mathrm{KL}}
+
\lambda\mathcal{L}_{\mathrm{CE}},
\label{eq:total-loss}
\end{equation}
where $\mathcal{L}_{\mathrm{CE}}$ supervises the final scam judgment.
For benign windows, no corresponding future scam progression exists from which to construct a privileged reflection, and we therefore use only $\mathcal{L}_{\mathrm{CE}}$.
This asymmetric objective follows the availability of privileged information: KL transfers additional anti-scam reasoning when such information exists, while CE preserves label-grounded discrimination for both scam and benign observations.

We first perform SFT on short scam and normal traces and then apply OPSD on long-horizon trajectories.
The large teacher, complete scam trace, and privileged reflection are used only during training; at inference time, ORACLE relies solely on the compact student and the observed $\tilde{\gW}_t$.

%% file: sec/exp.tex
\vspace{-4mm}
\section{Experiments}
\label{sec:experiments}
\vspace{-3mm}

\noindent\textbf{Implementation Details.}
Following the streaming evaluation protocol, each model processes a sliding window of size $W=10$ with a stride of 5.
We use Qwen3-235B via API calls during benchmark construction. Qwen3-32B is used as the teacher model for all the student models Qwen3-0.6B/1.7b/4B~\cite{yang2025qwen3}.
All experiments are conducted on 8$\times$\texttt{NVIDIA A6000} GPUs.
The proposed metrics Hit Rate (HR), Earliest Detection Position (EDP), False Alert Rate (FAR), and Partial-observation Alert Rate (PAR) are reported.
More configurations are provided in Appendix~\ref{app:implementation}.

\subsection{Main Results}

\noindent\textbf{Quantitative Analysis.}
Table~\ref{tab:main_results} compares \method{} with strong LLM and anti-scam
baselines under two settings.
The \textit{Streaming Cross-App} setting evaluates whether a model can issue
timely and reliable warnings from partially observed app-usage trajectories,
whereas the \textit{Single-App Content} setting follows conventional
content-level scam detection by merging complete event content into a single
input and measuring standard accuracy.

In the streaming setting, \method{} achieves the lowest EDP and highest PAR,
indicating earlier and more consistent recognition under partial observation.
Compared with GPT-5.1, it reduces EDP from 50.7 to 44.5 and improves PAR from
70.7 to 76.5, with a comparable FAR (14.8 vs.\ 13.9), although GPT-5.1 attains
a higher HR.
In the single-app setting, \method{} outperforms all
specialized anti-scam baselines.
These results highlight the advantage of cross-temporal reasoning when scam
evidence is incomplete and distributed across applications.

\input{tables/main_results}

\begin{figure*}[t]
    \centering
    \includegraphics[width=0.8\linewidth]{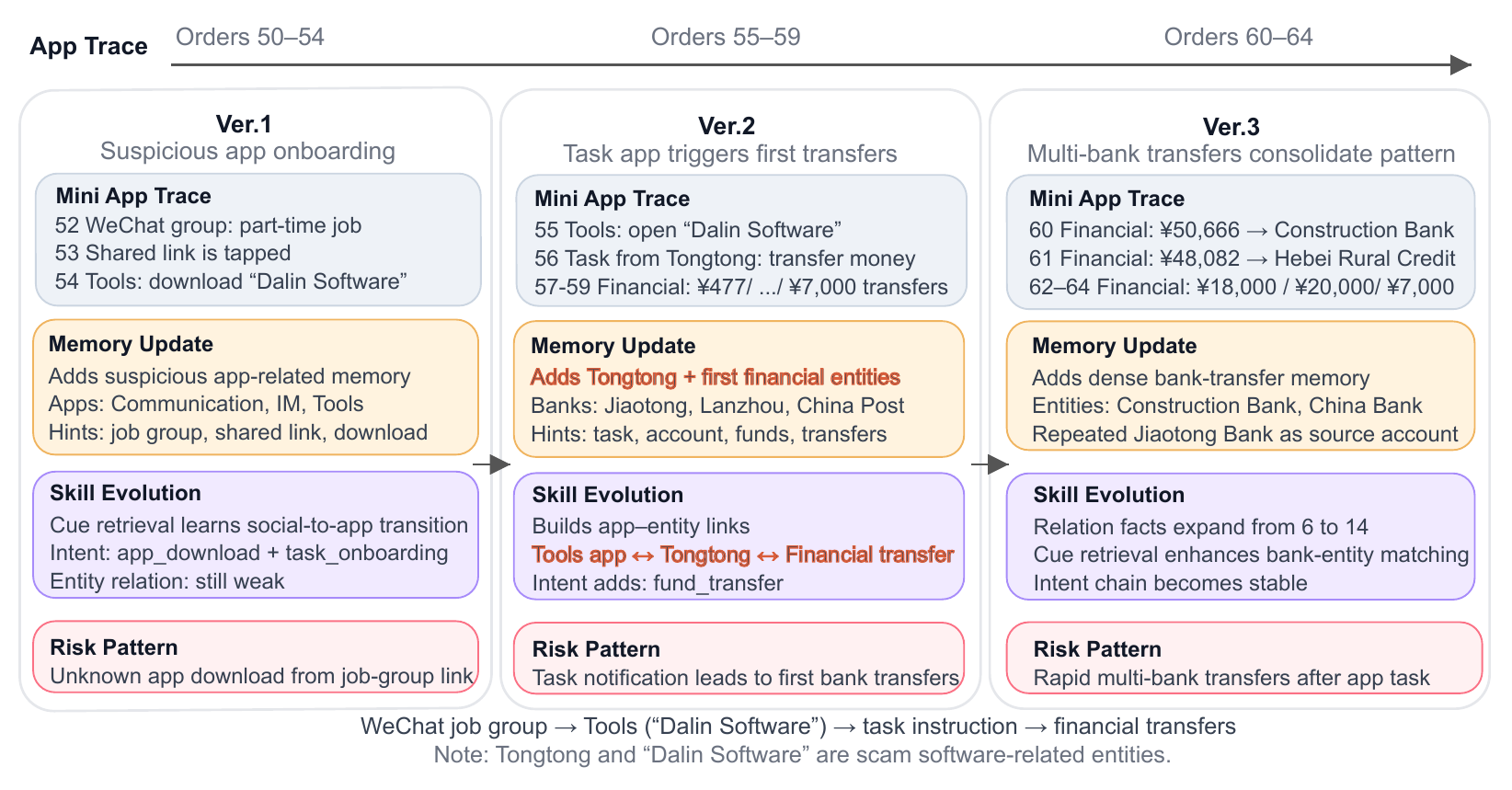}
    \caption{
    \textbf{Case-level memory--skill evolution.}
    ORACLE progressively links a job-group cue, installation of
    ``Dalin Software'', task instructions from ``Tongtong'', and repeated
    bank transfers into a reusable scam-stage skill.
    }
    \label{fig:memory_skill_evolution}
    \vspace{-3mm}
\end{figure*}

\noindent\textbf{Qualitative Analysis.}
Figure~\ref{fig:memory_skill_evolution} illustrates how evolving memory and
skills support long-horizon scam anticipation.
At an early stage, the trajectory contains only weak cues, including a
part-time-job group, a shared link, and installation of an external tool app.
Memory subsequently connects these observations with task instructions and
repeated financial transfers.
Meanwhile, skill evolution abstracts the trajectory into a reusable
progression:
\textit{job-group contact}
$\rightarrow$
\textit{third-party app onboarding}
$\rightarrow$
\textit{task instruction}
$\rightarrow$
\textit{fund transfer}.
This example illustrates the complementary roles of memory in preserving
cross-temporal evidence and skills in abstracting recurring behavioral
progressions for subsequent retrieval.

\subsection{Ablation Study}

\noindent\textbf{Effect of Evolving Context Management.}
Table~\ref{tab:ablation_evolving_context} progressively introduces the
components of the context manager while keeping the scam assessor fixed.
Person-centric memory substantially improves HR and PAR while reducing EDP,
showing the value of recovering historical evidence beyond the local window.
Static skill guidance further improves retrieval quality, reducing FAR from
21.1\% to 17.3\%.
With evolving skills, ORACLE achieves the best overall performance
(63.8\% HR, 44.5\% EDP, 14.8\% FAR, and 76.5\% PAR), indicating that memory
restores dispersed evidence while skill evolution progressively improves the
relevance of retrieved context.

\input{tables/ablation_agent}

\noindent\textbf{Comparison of Training Paradigms.}
We further compare different training paradigms under the same Qwen3-4B
backbone, training split, context setting, and streaming evaluation protocol.
As shown in Table~\ref{tab:matched_training}, standard SFT struggles to
effectively exploit long-horizon historical context, resulting in relatively
late detection and a high false-alert rate.
GRPO substantially improves this trade-off, while OPSD achieves the strongest
performance across all four metrics.
These results indicate that skill-guided on-policy distillation provides more
effective supervision for early scam anticipation than standard supervised or
reinforcement-based training.

\begin{table}[t]
\centering
\small
\caption{\textbf{Comparison of training paradigms.}
All settings use the same Qwen3-4B backbone, training split, context setting,
and streaming evaluation protocol.}
\label{tab:matched_training}
\begin{tabular}{lcccc}
\toprule
Training & HR$\uparrow$ & EDP$\downarrow$ &
FAR$\downarrow$ & PAR$\uparrow$ \\
\midrule
SFT
& 43.7 & 68.4 & 34.3 & 64.9 \\
GRPO
& 61.2 & 48.0 & 16.1 & 69.7 \\
OPSD
& \textbf{63.8} & \textbf{44.5} &
\textbf{14.8} & \textbf{76.5} \\
\bottomrule
\end{tabular}
\vspace{-2mm}
\end{table}

\noindent\textbf{Effect of OPSD Components.}
We further examine the contributions of skill-guided experience and training
objectives in Table~\ref{tab:opsd_internal_ablation}.
Under KL-only distillation, skill-guided experience improves HR, EDP, and PAR
over generic experience, with a particularly clear gain in PAR
(58.7\%$\rightarrow$65.9\%), indicating that scam-specific skills provide more
informative supervision for partial trajectories.
This advantage persists under KL+CE, where skill-guided experience further
improves EDP from 47.3\% to 44.5\% and PAR from 72.5\% to 76.5\%.
The full configuration achieves the strongest overall results, showing that
skill-guided privileged experience provides the key task-specific signal,
while CE complements it with better decision grounding and false-alert control.

\begin{table}[t]
\centering
\small
\caption{\textbf{Internal ablation of OPSD.}}
\label{tab:opsd_internal_ablation}
\begin{tabular}{llcccc}
\toprule
Experience & Objective &
HR$\uparrow$ & EDP$\downarrow$ & FAR$\downarrow$ & PAR$\uparrow$ \\
\midrule
Generic Exp. & KL
& 52.4 & 59.8 & 20.6 & 58.7 \\
Skill Exp. & KL
& 55.8 & 54.2 & 23.1 & 65.9 \\
Generic Exp. & KL + CE
& 62.6 & 47.3 & 15.3 & 72.5 \\
Skill Exp. & KL + CE
& \textbf{63.8} & \textbf{44.5} & \textbf{14.8} & \textbf{76.5} \\
\bottomrule
\end{tabular}
\vspace{-2mm}
\end{table}

\noindent\textbf{Reflection-Guided Correction.}
Figure~\ref{fig:teacher_student_reflection} presents a representative
false-negative case.
From the partial trajectory alone, the student interprets the observed behavior
as ordinary app usage.
The privileged reflection instead relates the suspicious communication to the
subsequent third-party app download, enabling the teacher to recognize the
emerging scam pattern.
This example illustrates how OPSD provides corrective supervision for weak
early-stage signals whose significance is not yet explicit from the partial
trajectory alone.

\begin{figure}[t]
    \centering
    \begin{tcolorbox}[
        width=\columnwidth,
        colback=gray!3,
        colframe=gray!35,
        boxrule=0.4pt,
        left=1mm,
        right=1mm,
        top=1mm,
        bottom=1mm
    ]
    \scriptsize\raggedright

    \textbf{Reflection.}
    The current window shows a progression from suspicious communication
    to a third-party app download, providing early scam-related evidence.

    \par\smallskip
    \textbf{Teacher.}

    {\ttfamily
    \{"reason": "The transition from an unknown part-time-job
    message to a third-party income-task app indicates scam-related
    behavior.", "fraud": true\}
    }

    \par\smallskip
    \textbf{Student.}

    {\ttfamily
    \{"reason": "Repeated app openings show normal usage without
    suspicious contacts or urgent requests.", "fraud": false\}
    }

    \end{tcolorbox}
    \vspace{-2mm}
    \caption{\textbf{Reflection-guided correction in OPSD.}
    Privileged anti-scam reflection provides corrective supervision for a
    false-negative student prediction.}
    \label{fig:teacher_student_reflection}
    \vspace{-6mm}
\end{figure}
\subsection{Temporal Pattern Analysis}

Our central hypothesis is that early scam anticipation requires recognizing
weak signals that are individually ambiguous but become informative when
accumulated along a behavioral trajectory.
We therefore examine whether reliable anticipation depends on cross-event
temporal context and the progression of scam-related behavior.

\noindent\textbf{Weakness of Isolated Signals.}
We first evaluate how informative individual events are without surrounding
trajectory context.
A TF--IDF classifier using only normalized event descriptions achieves 55.1\%
balanced accuracy, close to the 50\% random baseline.
This result indicates that isolated events provide only limited discriminative
information, as many scam-related cues remain weak and ambiguous when viewed
independently.
Reliable early anticipation therefore requires accumulating and jointly
interpreting such evidence across time.

\noindent\textbf{Dependency on Temporal Progression.}
We next examine whether ORACLE exploits how these weak signals evolve along the
trajectory.
We keep the trained model fixed and randomly shuffle the temporal order of
scam-related events while preserving their event content.
As shown in Table~\ref{tab:temporal_shuffle}, temporal perturbation reduces HR
from 63.8\% to 57.6\% and delays EDP from 44.5\% to 47.2\%.
More importantly, PAR decreases substantially from 76.5\% to 61.8\%.
This $14.7$-point drop indicates that disrupting the original behavioral
progression particularly weakens the model's ability to maintain scam
recognition across successive partial observations.
FAR also increases from 14.8\% to 16.3\%, suggesting that coherent temporal
context helps distinguish accumulating scam evidence from isolated suspicious
events.

The above results support that early anticipation cannot
be explained by isolated event-level signals alone.
ORACLE benefits from the temporal accumulation and progression of weak scam
evidence, which is precisely the information targeted by context management
and skill-guided distillation.

\begin{table}[t]
\centering
\small
\caption{\textbf{Effect of temporal-order perturbation.}
The content of scam-related events is preserved while their temporal order is
randomly shuffled, disrupting the original behavioral progression.}
\label{tab:temporal_shuffle}
\setlength{\tabcolsep}{4pt}
\begin{tabular}{lcccc}
\toprule
Setting & HR$\uparrow$ & EDP$\downarrow$ & FAR$\downarrow$ & PAR$\uparrow$ \\
\midrule
Original
& \textbf{63.8} & \textbf{44.5} & \textbf{14.8} & \textbf{76.5} \\
Temporal shuffle
& 57.6 & 47.2 & 16.3 & 61.8 \\
\bottomrule
\end{tabular}
\vspace{-2mm}
\end{table}
\subsection{Analysis}

\noindent\textbf{Generalization to Unseen Scams.}
Real-world scam strategies continuously evolve, making adaptation to previously unseen behavioral patterns important for practical deployment.
To evaluate this setting, we completely hold out three scam types
(E-commerce/Logistics Customer Service Scam, Fake Shopping or Service Scam,
and Fake In-game Transaction Scam).
We then retrain ORACLE using only the remaining seen scam types.
The alert threshold is calibrated exclusively on seen-type validation data and remains fixed.

Two OOD settings are compared.
In the \textit{OOD Frozen} setting, the original memory and skills remain unchanged.
In the \textit{OOD Evolving} setting, only the explicit memory and skill states are updated using adaptation cases.
For each held-out scam type, 20\% of the cases are reserved as an adaptation set and the remaining 80\% form an independent test set.
The adaptation set is used only in the \textit{OOD Evolving} setting, whereas both OOD settings are evaluated on the same 80\% test split.
No model parameters are updated in either setting.

As shown in Table~\ref{tab:ood_macro}, directly transferring the retrained model from seen to unseen scam types substantially degrades performance, with HR and PAR dropping from 64.2\% and 75.5\% to 48.7\% and 52.6\%, respectively, while EDP and FAR increase from 46.1\% and 13.9\% to 58.2\% and 22.4\%. Updating only the explicit memory and skill states partially recovers this gap without any parameter updates: \textit{OOD Evolving} improves HR to 56.4\% and PAR to 67.8\%, while reducing EDP to 51.3\% and FAR to 18.1\%. The particularly large recovery in PAR suggests that context evolution helps restore consistent recognition as unseen scam patterns unfold under partial observation, supporting the self-evolving context manager as a non-parametric adaptation mechanism that mitigates, though does not eliminate, distribution shift.

\begin{table}[t]
\centering
\small
\caption{\textbf{Generalization to unseen scam types.}
All results are obtained from a model retrained with three scam types held out.}
\label{tab:ood_macro}
\setlength{\tabcolsep}{4pt}
\begin{tabular}{lcccc}
\toprule
Setting & HR$\uparrow$ & EDP$\downarrow$ & FAR$\downarrow$ & PAR$\uparrow$ \\
\midrule
Seen-type ID
& \textbf{64.2} & \textbf{46.1} & \textbf{13.9} & \textbf{75.5} \\
OOD Frozen
& 48.7 & 58.2 & 22.4 & 52.6 \\
OOD Evolving
& 56.4 & 51.3 & 18.1 & 67.8 \\
\bottomrule
\end{tabular}
\vspace{-2mm}
\end{table}

\noindent\textbf{Failure Analysis.}
We identify three representative failure modes.
First, when early observations contain only generic social or financial actions
without sufficiently distinctive entity-level cues, ORACLE may delay its first
warning until stronger evidence emerges.
Second, benign behavioral sequences that resemble common scam progressions,
such as interactions with unfamiliar users followed by financial-app usage,
can still trigger false alerts.
Third, entity-centric retrieval becomes less reliable when the same entity
appears across many unrelated interactions, causing irrelevant historical
events to enter the augmented context.
These failures highlight two remaining challenges for early scam anticipation:
distinguishing genuinely informative weak signals from benign behavioral
coincidences, and suppressing retrieval noise as long-term memory grows.
Improved uncertainty modeling and relevance-aware memory consolidation may
further address these limitations.

%% file: tables/main_results.tex
\begin{table*}[t]
\small
\centering
\caption{
\textbf{Main results on the scam anticipation benchmark.}
Best results are shown in \textbf{bold}. N/A indicates that the method is not applicable to the streaming cross-app setting.
}
\label{tab:main_results}
\begin{tabular}{lccccc}
\toprule
\textbf{Method}
& \multicolumn{4}{c}{\textbf{Streaming Cross-App}}
& \textbf{Single-App Content} \\
\cmidrule(lr){2-5} \cmidrule(lr){6-6}
& \textbf{HR(\%)$\uparrow$}
& \textbf{EDP(\%)$\downarrow$}
& \textbf{FAR(\%)$\downarrow$}
& \textbf{PAR(\%)$\uparrow$}
& \textbf{Acc.(\%)$\uparrow$} \\
\midrule

\multicolumn{6}{c}{\textit{General LLMs}} \\
GPT-5.1~\cite{openai2025gpt5} & 73.4 & 50.7 & 13.9 & 70.7 & 91.5 \\
Claude-4-Sonnet~\cite{anthropic2025claude-sonnet4} & 71.6 & 53.4 & 13.3 & 66.2 & 91.6 \\
Gemini-3-Flash~\cite{google2025gemini3} & 71.1 & 54.5 & 14.8 & 61.7 & 91.1 \\
Kimi-2.5~\cite{team2026kimi} & 70.2 & 55.7 & 13.0 & 62.8 & 90.9 \\
GLM-4.6~\cite{zeng2025glm} & 69.8 & 57.0 & 11.7 & 64.1 & 90.8 \\
DeepSeek-V4~\cite{deepseekv4pro2026} & 69.1 & 46.3 & 12.3 & 62.0 & 89.4 \\
Grok-4~\cite{xai2025grok4} & 67.0 & 60.6 & 15.1 & 59.0 & 90.1 \\
\midrule

\multicolumn{6}{c}{\textit{Specialized Methods (Qwen3-4B)}} \\
BotWar~\cite{basta2025bot}        & N/A & N/A & N/A & N/A & 88.7 \\
ScamGPT-J~\cite{tan2024scamgpt}   & N/A & N/A & N/A & N/A & 89.4 \\
AntiFraud~\cite{ma2025teleantifraud} & N/A & N/A & N/A & N/A & 89.1 \\
FraudR1~\cite{fraudr1}            & N/A & N/A & N/A & N/A & 90.2 \\
SafeQAQ~\cite{safeqaq}            & N/A & N/A & N/A & N/A & 89.8 \\
\midrule

\method{} (Qwen3-0.6B) & 57.8 & 59.8 & 26.7 & 71.0 & 87.9 \\
\method{} (Qwen3-1.7B) & 62.9 & 47.6 & 18.9 & 74.0 & 89.1 \\
\method{} (Qwen3-4B) & \textbf{63.8} & \textbf{44.5} & \textbf{14.8} & \textbf{76.5} & \textbf{91.4} \\

\bottomrule
\end{tabular}
\vspace{-2mm}
\end{table*}

%% file: tables/ablation_agent.tex
\begin{table*}[!t]
\small
\centering
\caption{\textbf{Ablation study of evolving context management.}
All settings use the same scam assessor.}
\label{tab:ablation_evolving_context}
\begin{tabular}{ccc|cccc}
\toprule
\multicolumn{3}{c|}{\textbf{Context Management}} &
\multicolumn{4}{c}{\textbf{Performance}} \\
\cmidrule(lr){1-3}
\cmidrule(lr){4-7}
\textbf{Memory} & \textbf{Static Skill} & \textbf{Evolving Skill}
& \textbf{HR(\%)$\uparrow$}
& \textbf{EDP(\%)$\downarrow$}
& \textbf{FAR(\%)$\downarrow$}
& \textbf{PAR(\%)$\uparrow$} \\
\midrule
--         & --         & --         & 48.6 & 61.2 & 22.5 & 51.3 \\
\checkmark & --         & --         & 56.9 & 53.8 & 21.1 & 64.8 \\
\checkmark & \checkmark & --         & 60.2 & 49.1 & 17.3 & 70.6 \\
\checkmark & \checkmark & \checkmark
           & \textbf{63.8} & \textbf{44.5}
           & \textbf{14.8} & \textbf{76.5} \\
\bottomrule
\end{tabular}
\vspace{-5mm}
\end{table*}


%% file: sec/conclusion.tex
\section{Conclusion}

We study early scam anticipation from streaming app-usage trajectories, where scam evidence is fragmented across time and applications.
We introduce a long-horizon benchmark for evaluating timely and reliable scam anticipation and propose ORACLE, which combines a self-evolving context manager with On-Policy Skill-guided Distillation.
Experiments show that ORACLE improves early and consistent scam anticipation
while maintaining competitive false-alert control, highlighting the importance of cross-temporal reasoning over partial behavioral trajectories for scam intervention.

%% file: sec/appendix.tex
\appendix
\section{Broader Impacts}

This work aims to support earlier and more reliable scam intervention from streaming app-usage trajectories. Compared with existing single-app content-based scam detection methods, our framework operates on structured app-level events and short summarized content rather than full raw messages or detailed interaction logs. This design reduces the amount of sensitive information required for detection, while still enabling cross-app temporal reasoning. As a result, the system can provide earlier and more consistent scam warnings with a more privacy-conscious representation.

Despite this reduced information footprint, the approach still inevitably relies on partial user behavior signals, which raises privacy concerns if deployed without appropriate safeguards. Continuous monitoring of app-usage patterns, even at a summarized level, may expose sensitive behavioral traits. In addition, incorrect predictions may lead to unintended consequences: false alerts could disrupt benign activities or reduce user trust, while delayed or missed alerts may fail to prevent scam-related harm. The reliance on memory retrieval may also introduce noisy or outdated context, potentially affecting decision reliability. Finally, the system could be misused for large-scale behavioral monitoring or profiling beyond fraud prevention.

These considerations highlight the importance of responsible deployment. Practical use should incorporate privacy-preserving mechanisms (e.g., local processing and minimal data retention), transparent system design, confidence-aware alerting, and safeguards against misuse, such as access control and auditing.
\section{Related Works}
\subsection{Scam Anticipation}
Recent studies have begun to explore automatic scam detection for smartphones. 
Bot-Wars~\cite{basta2025bot} counters phone scams through a two-layer prompt architecture and dual-stream chain-of-thought reasoning, enabling LLM-driven scam-baiting agents to generate demographically realistic victim personas and adversarial strategies. 
ScamGPT-J~\cite{tan2024scamgpt} addresses instant-messaging scam detection with a fine-tuned large language model that simulates scammer responses in real time, helping users identify suspicious interactions through analogy-based reasoning. 
Both TeleAntiFraud~\cite{ma2025teleantifraud} and SafeQAQ~\cite{safeqaq} address voice-based telecom fraud analysis from telephone audio by proposing slow-thinking multimodal frameworks. 
Fraud-R1~\cite{fraudr1} assesses LLM robustness against escalating fraud inducements from credibility building to emotional manipulation, across five real-world scam types.

Although promising, these studies mainly focus on single applications or isolated interactions, such as one phone call or one message thread. 
They are therefore less suited to long-horizon cross-app scams, where fraudulent intent emerges gradually across multiple contacts and actions. 
Moreover, their reliance on detailed conversational content raises privacy concerns for continuous smartphone monitoring, leaving a gap in privacy-aware scam detection from long-horizon app-usage histories.
\subsection{Knowledge Distillation}
OPCD~\cite{opcd} and Self-Distilled Reasoner~\cite{zhao2026self} present a framework that internalizes contextual knowledge by training student models on their own generated trajectories while minimizing reverse KL divergence to a context-conditioned teacher. However, the contextual knowledge it exploits is mainly derived from relatively simple raw environmental feedback, such as observations and rewards from text-game environments. OEL~\cite{oel} extends on-policy distillation by introducing an online learning loop that extracts and consolidates experiential knowledge from real-world deployment trajectories. Unlike approaches that rely on simple immediate observations, the contextual knowledge it leverages is recursively accumulated from long-horizon multi-turn interactions and requires sophisticated reasoning-based extraction to produce transferable and structured insights.


\section{Benchmark Construction Details}
\label{app:benchmark}

\begin{figure*}[t]
    \centering
    \includegraphics[width=0.99\linewidth]{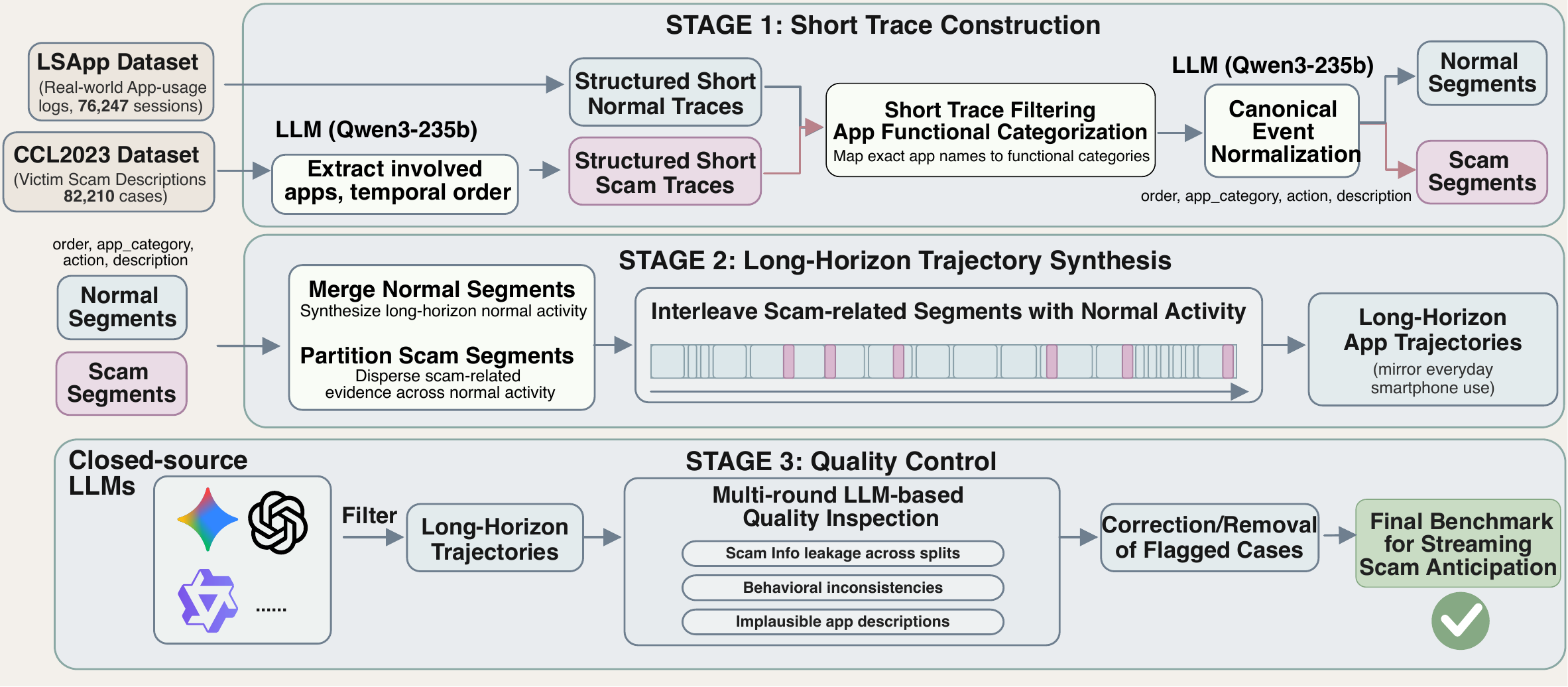}
    \caption{
    \textbf{Dataset curation pipeline for streaming scam anticipation.}
    The benchmark is constructed in three stages:
    (\textbf{1}) heterogeneous source records are converted into canonical
    normal and scam short traces;
    (\textbf{2}) normal traces are merged into extended activity histories and
    temporally ordered scam-related segments are interleaved with normal
    activity to form long-horizon trajectories;
    and (\textbf{3}) synthesized trajectories undergo quality inspection and
    correction or removal of flagged cases.
    }
    \label{fig:data_curation}
\end{figure*}

Figure~\ref{fig:data_curation} summarizes our three-stage benchmark construction
pipeline.
\textbf{Stage 1: Short Trace Construction} converts heterogeneous source data
into canonical normal and scam short traces with a unified event representation
and textual granularity.
\textbf{Stage 2: Long-Horizon Trajectory Synthesis} concatenates normal traces
to form extended everyday activity and interleaves temporally ordered
scam-related segments within this background, producing long-horizon trajectories
with progressively emerging scam evidence.
\textbf{Stage 3: Quality Control} inspects the synthesized trajectories for
semantic and temporal inconsistencies and corrects or removes flagged cases.
Together, these stages transform real-world normal app usage and victim-reported
scam processes into a benchmark for streaming scam anticipation.

\subsection{Unified Short Trace Construction}

Our benchmark is constructed primarily from CCL2023~\cite{sun-etal-2023-ccl23}
and LSApp~\cite{lsapp}, which provide complementary representations of scam
behavior and everyday smartphone usage.

CCL2023 contains detailed natural-language police-report narratives from scam
victims.
These records describe the progression of complete scam processes and typically
include the applications involved, temporal information, anonymized interacting
persons, user operations, and conversational context.
We use Qwen3-235B to parse each case according to its temporal progression and
extract a sequence of application interactions, user actions, interacting
entities, and operation or conversation descriptions.

LSApp provides structured real-world app-usage logs.
Each event records the application name, temporal order, and one of three
coarse user-action states:
\texttt{Open}, \texttt{Close}, or \texttt{Interaction}.
We use this representation as the basis of a unified event schema:
\[
(\texttt{order},\ \texttt{app\_category},\
\texttt{action},\ \texttt{description}).
\]

For both sources, exact application names are mapped to 12 functional
categories.
CCL2023 operations are additionally mapped to the three LSApp action states,
and the extracted scam events are reformatted into the same schema.
Short traces are filtered before subsequent synthesis to retain valid event
sequences suitable for benchmark construction.

For normal events requiring additional interaction-level textual context,
LOCOMO~\cite{locomo} is used only as an auxiliary source of
coarse person-centric content rather than directly transferring its original
memory-style summaries.
Textual content derived from CCL2023 and LOCOMO is re-summarized using the same
LLM, prompt template, output schema, and length constraint.
Person identifiers follow the same anonymization scheme, and exact application
names are represented only by their functional categories.

This stage produces canonical short normal and scam traces with consistent
event structure and comparable textual granularity, which are subsequently
used for long-horizon trajectory synthesis.

\subsection{Long-Horizon Trajectory Synthesis}

Normal short traces are concatenated to construct extended everyday app-usage
histories.
Each scam trace is partitioned into multiple temporally ordered sub-segments
following the behavioral progression described in the original victim record.
These scam-related sub-segments are distributed across the normal activity
stream, with normal events inserted between them, while preserving the original
relative order among scam-related events.
Thus, scam evidence emerges progressively throughout the trajectory rather
than appearing as a single contiguous malicious block.

We refer to the interval from the first scam-related event to the last
scam-related event as the \emph{scam process interval} $[s,e]$.
Normal activities may occur within this interval, reflecting realistic
interleaving between scam-related and everyday smartphone behavior.

The resulting trajectories consistently follow the representation
\[
\texttt{order}, 
\texttt{app\_category}, 
\texttt{action}, 
\texttt{description}.
\]

\subsection{Quality Control}

We perform semantic quality control using LLM judges from different model
families, including GPT, Qwen, and Gemini.
The judges inspect temporal inconsistencies, implausible application
transitions, mismatches between actions and descriptions, and malformed events.
Flagged cases are corrected or removed.

\subsection{Benchmark Statistics}

\begin{figure*}[t]
    \centering
    \begin{subfigure}[t]{0.4\textwidth}
        \centering
        \includegraphics[width=\linewidth]
        {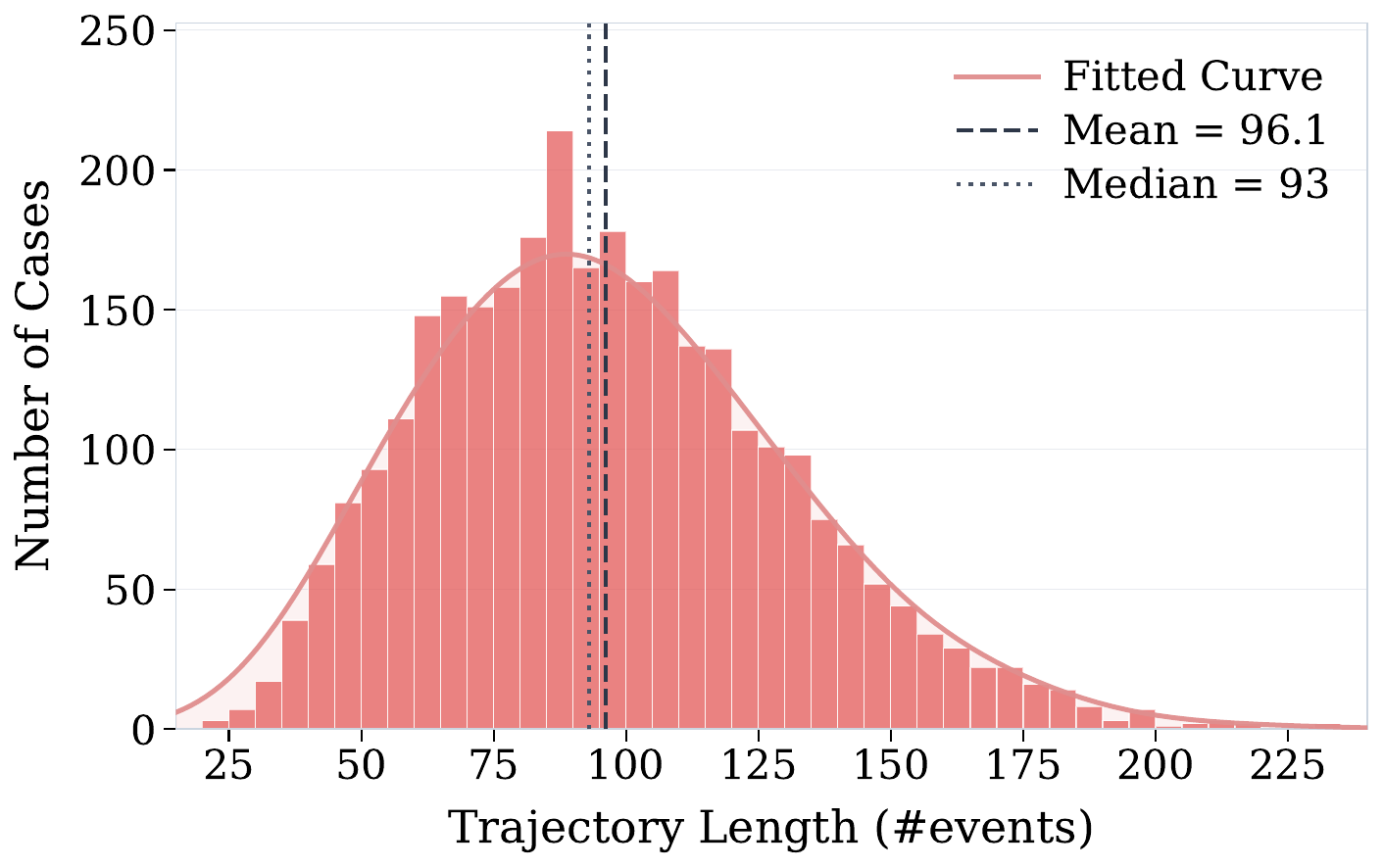}
        \caption{Trajectory-length distribution.}
        \label{fig:trajectory_length}
    \end{subfigure}
    \hfill
    \begin{subfigure}[t]{0.55\textwidth}
        \centering
        \includegraphics[width=\linewidth]
        {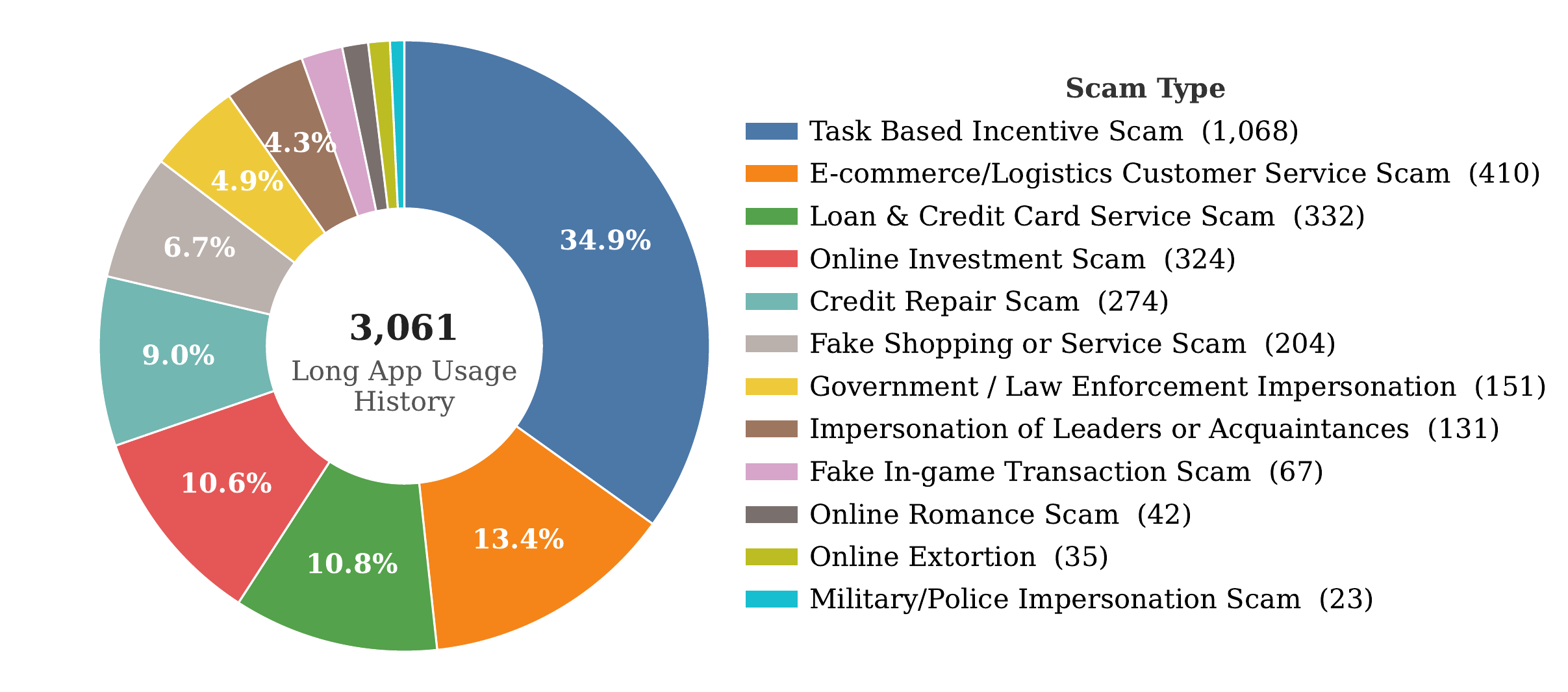}
        \caption{Distribution of scam types.}
        \label{fig:scam_type_distribution}
    \end{subfigure}
    \caption{
    \textbf{Benchmark statistics.}
    (\textbf{a}) Distribution of long-horizon trajectory lengths
    (mean: 96.1 events; median: 93 events).
    (\textbf{b}) Distribution of 12 scam types across the 3,061 long-horizon
    trajectories.
    }
    \label{fig:data_statistics}
\end{figure*}

The curated benchmark contains 57,662 short app-usage traces, which are
aggregated into 3,061 long-horizon trajectories covering 12 scam types.
Figure~\ref{fig:data_statistics} summarizes the trajectory-length and scam-type
distributions.
The long trajectories contain an average of 96.1 events, with a median length
of 93 events, substantially exceeding the local observation window used in
streaming evaluation.

We further report the distribution of scam-related lengths in
Table~\ref{tab:scam_length}.
The number of scam-related events ranges from 5 to 18, with a median of 7.
These statistics characterize the varying amount of scam-related information
available across trajectories and motivate explicitly accounting for different
scam-process durations in streaming evaluation.

\begin{table*}[t]
\centering
\small
\caption{
\textbf{Distribution of scam-related lengths.}
The length denotes the number of scam-related events associated with each
trajectory.
}
\label{tab:scam_length}
\setlength{\tabcolsep}{5pt}
\begin{tabular}{lcccccccccc}
\toprule
Percentile
& 1\% & 5\% & 10\% & 25\% & 50\%
& 75\% & 90\% & 95\% & 99\% & 100\% \\
\midrule
Scam-related length
& 5 & 5 & 5 & 5 & 7 & 10 & 13 & 15 & 18 & 18 \\
\bottomrule
\end{tabular}
\end{table*}


\section{Evaluation Metric Definitions}
\label{app:metrics}

Let a trajectory $\mathcal{T}$ of length $L$ contain a scam process interval
$[s,e]$, where $0\leq s\leq e<L$.
The interval begins at the first scam-related event and ends at the last
scam-related event, and may contain interleaved normal activities.
At timestep $t$, the system observes a window
$[s_w(t),e_w(t)]$ of length $W$.

The assessor internally outputs
\[
z_t\in
\{\textsc{Normal},\textsc{Risky},\textsc{Scam}\}.
\]
For evaluation, we use the binary prediction
\[
\hat y_t
=
\mathbf{1}
[z_t\in\{\textsc{Risky},\textsc{Scam}\}],
\]
such that both \textsc{Risky} and \textsc{Scam} are treated as positive scam
predictions.

\noindent\textbf{Hit Rate and False Alert Rate.}
For trajectory $i$, let
\[
\mathcal{T}^{(i)}_{\mathrm{scam}}
=
\{t\mid e_w(t)\in[s_i,e_i]\},
\]
and let
\[
\mathcal{T}^{(i)}_{\mathrm{out}}
=
\{t\mid e_w(t)\notin[s_i,e_i]\}.
\]
For $N$ scam trajectories, HR is defined as
\begin{equation}
\mathrm{HR}
=
\frac{1}{N}
\sum_{i=1}^{N}
\mathbf{1}
\left[
\exists t\in\mathcal{T}^{(i)}_{\mathrm{scam}}
:\hat y_t=1
\right].
\end{equation}
Thus, HR measures whether each scam trajectory is successfully detected at
least once while the observation window ends within the scam process.

FAR is computed at the window level:
\begin{equation}
\mathrm{FAR}
=
\frac{
\sum_{i=1}^{N}
\sum_{t\in\mathcal{T}^{(i)}_{\mathrm{out}}}
\mathbf{1}[\hat y_t=1]
}{
\sum_{i=1}^{N}
|\mathcal{T}^{(i)}_{\mathrm{out}}|
}.
\end{equation}
It therefore measures spurious positive predictions for windows whose
endpoints fall outside the scam process interval.

\noindent\textbf{Earliest Detection Position.}
We define the valid detection set as
\[
\mathcal{T}_{\mathrm{valid}}
=
\left\{
t\mid
\hat y_t=1,\;
e_w(t)\in[s,e]
\right\}.
\]
For each valid detection, its normalized position is
\[
p_t
=
\frac{e_w(t)-s}{e-s+1}.
\]
EDP is defined as
\begin{equation}
\mathrm{EDP}
=
\begin{cases}
\min\limits_{t\in\mathcal{T}_{\mathrm{valid}}} p_t,
& \mathcal{T}_{\mathrm{valid}}\neq\emptyset,\\
1,
& \text{otherwise}.
\end{cases}
\end{equation}
A smaller EDP indicates earlier anticipation; trajectories without a valid
detection are assigned $\mathrm{EDP}=1$.
The reported EDP is averaged across test trajectories.

\noindent\textbf{Partial-observation Alert Rate.}
Let
\[
S=e-s+1
\]
denote the length of the scam process interval.
For each window ending within $[s,e]$, we define its observable scam-process
coverage as
\begin{equation}
c_t
=
\frac{
|[s_w(t),e_w(t)]\cap[s,e]|
}{
\min(W,S)
}.
\end{equation}
The denominator $\min(W,S)$ corresponds to the maximum portion of the scam
process interval that can be observed within a single window and accommodates
processes both shorter and longer than $W$.

We define the partial-observation candidate set as
\begin{equation}
\mathcal{T}_{\mathrm{pre}}
=
\left\{
t\mid
e_w(t)\in[s,e],\;
c_t\geq0.5
\right\}.
\end{equation}
PAR for each eligible trajectory is
\begin{equation}
\mathrm{PAR}
=
\frac{
\sum_{t\in\mathcal{T}_{\mathrm{pre}}}
\mathbf{1}[\hat y_t=1]
}{
|\mathcal{T}_{\mathrm{pre}}|
},
\end{equation}
for $|\mathcal{T}_{\mathrm{pre}}|>0$.
The reported PAR is averaged across eligible trajectories.

EDP and PAR capture complementary temporal properties:
EDP measures the position of the \emph{first} successful anticipation,
whereas PAR measures how \emph{consistently} the model maintains positive
recognition under partial observation.


\section{Implementation Details}
\label{app:implementation}

\subsection{Streaming Evaluation}

All models follow the same online evaluation protocol.
A sliding window of size $W=10$ is applied with a stride of 5.
At each step, the model receives the current observation window together with
historical information retrieved by the context manager.
The alert threshold is calibrated using the validation split and remains fixed
during test-time evaluation.

\subsection{Benchmark Processing}

Qwen3-235B is used during benchmark construction to parse CCL2023 narratives
and extract the applications, temporal progression, user operations, and
scam-related conversational information required for structured trace
construction.

The screen analyzer uses Qwen3-8B-VL to extract entities and concise
event-level summaries from incoming app events.

\subsection{Supervised Fine-tuning}

We use Qwen3-4B~\cite{yang2025qwen3} as the default scam assessor.
SFT is performed on short scam and normal traces for two epochs using a
learning rate of $1\times10^{-5}$.
The per-device batch size is 4 with 2 gradient-accumulation steps, resulting
in an effective global batch size of 64 across 8 GPUs.

\subsection{On-Policy Skill-guided Distillation}

Starting from the SFT checkpoint, OPSD is conducted for one additional epoch
on long-horizon trajectories.
The Qwen3-4B assessor serves as the student, while a larger frozen teacher 
provides privileged reflection-conditioned supervision during training.
The teacher is used only during OPSD and is discarded at inference.

We use a learning rate of $5\times10^{-6}$, a CE-loss weight of
$\lambda=0.1$, GPU memory utilization of 0.4, and a global training batch size
of 8.
All student training experiments are conducted on
8$\times$\texttt{NVIDIA A6000} GPUs.
The complete OPSD training process takes approximately 7.5 hours.


\section{Additional Details of the Self-Evolving Context Manager}
\label{app:context-manager}

The Screen Analyzer $\phi_{\mathrm{scr}}$ maps each event $\rvh_t$ to extracted
entities $\gE_t$, such as person names or phone numbers, and a concise content
summary $\rvs_t$:
\[
\phi_{\mathrm{scr}}(\rvh_t)
\mapsto
(\gE_t,\rvs_t).
\]

These observations are stored in a person-centric memory $\gM_t$, which
organizes interactions by entity in chronological order:
\[
\gM_t(e)
=
\{(a_i,\rvs_i,\sigma_i)\}_{i=1}^{n_e},
\]
where $\sigma_i$ denotes global temporal order.

Memory is updated incrementally as
\begin{equation}
\gM_{t+1}
=
\textsc{Update}
(\gM_t,\phi_{\mathrm{scr}}(\rvh_{t+1})).
\end{equation}

At inference time, entities appearing in the current observation window
$\gW_t$ are used to retrieve associated interactions from $\gM_t$.
Retrieval is governed by the skill library $\gS_t$, where each skill encodes
a high-level anti-scam heuristic, such as prioritizing recent financial-app
events involving the same entity.

The system evolves at two timescales.
Memory is updated with every new event, while the skill library is updated
less frequently when the assessor produces a \textsc{Risky} or \textsc{Scam}
judgment.
Given rationale $R_t$, augmented context $\tilde{\gW}_t$, and internal
judgment $z_t$, the skill state evolves as
\begin{equation}
\gS_{t+1}
=
\textsc{Evolve}
(\gS_t,R_t,\tilde{\gW}_t,z_t).
\end{equation}

The scam-risk assessor outputs a rationale $R_t$, an internal risk state
$z_t$, and scam probability
\[
q_t
=
\pi_\theta(S=1\mid\tilde{\gW}_t).
\]
An alert is raised when $q_t>\tau$, where $\tau$ is calibrated on the
validation split.
The base model parameters remain frozen during online memory and skill
evolution.


\section{On-Policy Skill-guided Distillation Details}
\label{app:opsd}

\subsection{Privileged Reflection Construction}

For a scam-related training window $\tilde{\gW}_t$ that partially observes a
scam process, the student $\pi_{\theta}$ receives only the augmented partial
trajectory, matching the information available at inference time.
A stronger frozen teacher $\pi_{\mathrm{T}}$ additionally receives a privileged
anti-scam reflection $E_t$.

The reflection is constructed by comparing the currently visible partial trace
with the corresponding complete scam trace under a scam-type-specific skill.
It summarizes how currently weak or ambiguous cues may evolve into later
explicit scam behavior, thereby converting future scam progression into
training-time supervision.

OPSD performs supervision on token trajectories generated by the current
student policy.
The teacher evaluates the same student-visited trajectory while additionally
conditioning on $E_t$, aligning supervision with states encountered by the
student rather than relying solely on fixed offline demonstrations.

\subsection{Token-Space Distillation}

Distillation is performed directly in autoregressive token space without a
separate classification head.
The model generates a structured output containing both a natural-language
rationale and a final risk judgment.

The KL objective aligns the student distribution with the privileged teacher:
\begin{equation}
\mathcal{L}_{\mathrm{KL}}
=
D_{\mathrm{KL}}\!\left(
\pi_{\theta}(\cdot\mid\tilde{\gW})
\;\|\;
\pi_{\mathrm{T}}(\cdot\mid\tilde{\gW},E)
\right).
\end{equation}

We use reverse KL to encourage the student to concentrate on high-probability
reasoning paths identified by the privileged teacher rather than forcing it to
cover low-probability alternatives~\cite{zhao2026self}.

Privileged anti-scam reflections are generated only for scam-related windows.
For benign windows, no corresponding future scam progression exists and the
objective reduces to standard cross-entropy:
\[
\mathcal{L}
=
\mathcal{L}_{\mathrm{CE}}.
\]

For scam-related windows, the objective becomes
\[
\mathcal{L}
=
\mathcal{L}_{\mathrm{KL}}
+
\lambda\mathcal{L}_{\mathrm{CE}},
\]
where $\lambda>0$ controls the contribution of grounded label supervision.
The KL term transfers reflection-informed reasoning, whereas CE anchors the
final judgment to the ground-truth scam label and helps control false alerts.

\subsection{Training Process}

The scam assessor is trained in two stages.
First, supervised fine-tuning is performed on short scam and normal traces to
establish basic risk recognition and the required output format.

Second, OPSD is applied to long-horizon trajectories.
For scam-related windows, the corresponding complete scam trace is used only
during training to construct the privileged reflection $E_t$.
The student is optimized from the partial augmented observation
$\tilde{\gW}_t$, while the frozen teacher additionally receives $E_t$ for
computing the token-level KL term.

At inference time, neither the complete future scam trace nor the privileged
reflection or teacher model is available.
ORACLE operates solely from the compact student, the observed streaming
trajectory, and retrieved historical context.

\section{OOD Generalization Details}
\label{app:ood}

\subsection{Experimental Protocol}

We evaluate distribution shift by completely holding out three scam types:
\begin{itemize}
    \item E-commerce/Logistics Customer Service Scam;
    \item Fake Shopping or Service Scam;
    \item Fake In-game Transaction Scam.
\end{itemize}

These types contain 410, 204, and 67 cases, respectively.
All corresponding examples are excluded from SFT, OPSD, initial memory
construction, initial skill construction, validation, and threshold calibration.
ORACLE is retrained using only the remaining seen scam types.

For each held-out type, 20\% of cases are reserved as an adaptation set and
the remaining 80\% form an independent test set.
This gives adaptation/test sizes of 82/328, 41/163, and 13/54 for the three
types, respectively.

We compare two settings:
\begin{itemize}
    \item \textbf{OOD Frozen}: the original memory and skill states remain
    unchanged;
    \item \textbf{OOD Evolving}: only explicit memory and skill states are
    updated using the corresponding adaptation cases.
\end{itemize}

The adaptation set is used only in the \textit{OOD Evolving} setting, whereas
both settings are evaluated on the same 80\% test split.
The alert threshold is calibrated only on seen-type validation data and remains
fixed during OOD evaluation.
No model parameters are updated in either setting.

\subsection{Transfer to Individually Held-out Scam Types}
\label{app:cross_type_transfer}

We further examine whether behavioral knowledge learned from existing scam
categories can transfer to a previously unseen scam type.
This experiment differs from the joint OOD evaluation in the main paper:
rather than simultaneously excluding all three target scam types, we conduct
three independent leave-one-type-out experiments.
In each experiment, only the target scam type is excluded from SFT, OPSD,
initial memory and skill construction, validation, and threshold calibration,
while all remaining scam types are retained during training.

For each target type, 20\% of its cases are reserved for non-parametric
adaptation and the remaining 80\% form the test set.
The adaptation cases are used only in the \textit{Evolving} setting.
In the \textit{Frozen} setting, the trained model and its original memory and
skill states are directly evaluated on the target type.
In the \textit{Evolving} setting, only the explicit memory and skill states are
updated using the adaptation cases.
The model parameters and alert threshold remain fixed in both settings.

\begin{table*}[t]
\centering
\small
\caption{\textbf{Transfer to individually held-out scam types.}
Each target scam type is evaluated in a separate leave-one-type-out experiment,
while all other scam types remain available during training.
Only memory and skill states are updated in the \textit{Evolving} setting;
model parameters and the alert threshold remain fixed.}
\label{tab:ood_per_type}
\begin{tabular}{llcccc}
\toprule
Target Scam Type & Setting &
HR$\uparrow$ & EDP$\downarrow$ &
FAR$\downarrow$ & PAR$\uparrow$ \\
\midrule

E-commerce/Logistics
& Frozen
& 51.2 & 56.8 & 21.3 & 55.6 \\
& Evolving
& \textbf{59.4} & \textbf{49.8} &
\textbf{17.2} & \textbf{69.5} \\

\midrule

Fake Shopping/Service
& Frozen
& 49.0 & 58.1 & 22.5 & 53.1 \\
& Evolving
& \textbf{57.8} & \textbf{50.7} &
\textbf{18.0} & \textbf{67.9} \\

\midrule

Fake In-game Transaction
& Frozen
& 45.9 & 61.0 & 24.6 & 48.7 \\
& Evolving
& \textbf{51.9} & \textbf{56.2} &
\textbf{21.4} & \textbf{60.1} \\

\bottomrule
\end{tabular}
\end{table*}

As shown in Table~\ref{tab:ood_per_type}, context evolution improves all four
metrics for each individually held-out target type.
The gains are particularly pronounced in PAR, which increases from
55.6\% to 69.5\% for E-commerce/Logistics, from 53.1\% to 67.9\% for
Fake Shopping/Service, and from 48.7\% to 60.1\% for Fake In-game Transaction.
HR and EDP also consistently improve, while FAR decreases across all three
target types.

Because the remaining scam categories are still available during training,
these experiments evaluate whether ORACLE can transfer previously learned
behavioral patterns to a new scam category and further adapt that knowledge
through explicit memory and skill evolution.
The consistent improvements from \textit{Frozen} to \textit{Evolving} suggest
that ORACLE benefits from both transferable cross-type knowledge and newly
observed target-type experience without updating model parameters.

\begin{table*}[!hbpt]
\centering
\caption{\textbf{Sensitivity analysis of CE-loss weight and window length.}
Best results are shown in \textbf{bold}.}
\label{tab:sensitivity_ce_window}
\setlength{\tabcolsep}{5pt}
\small

\begin{minipage}[t]{0.48\textwidth}
\centering
\caption*{\textbf{(a) CE-Loss Weight}}
\vspace{0.25em}
\begin{tabular}{lcccc}
\toprule
\textbf{CE Weight} &
\textbf{HR$\uparrow$} &
\textbf{EDP$\downarrow$} &
\textbf{FAR$\downarrow$} &
\textbf{PAR$\uparrow$} \\
\midrule
0            & 55.8 & 54.2 & 23.1 & 65.9 \\
0.01         & 62.1 & 45.7 & 16.9 & 71.6 \\
\textbf{0.1} & \textbf{63.8} & \textbf{44.5}
             & \textbf{14.8} & \textbf{76.5} \\
1.0          & 61.1 & 47.2 & 16.4 & 72.8 \\
\bottomrule
\end{tabular}
\end{minipage}
\hfill
\begin{minipage}[t]{0.48\textwidth}
\centering
\caption*{\textbf{(b) Window Length}}
\vspace{0.25em}
\begin{tabular}{lcccc}
\toprule
\textbf{Window} &
\textbf{HR$\uparrow$} &
\textbf{EDP$\downarrow$} &
\textbf{FAR$\downarrow$} &
\textbf{PAR$\uparrow$} \\
\midrule
1           & 49.2 & 60.3 & 23.7 & 53.0 \\
5           & 60.6 & 48.1 & 17.4 & 70.3 \\
\textbf{10} & \textbf{63.8} & \textbf{44.5}
            & \textbf{14.8} & \textbf{76.5} \\
15          & 62.1 & 46.0 & 15.6 & 74.2 \\
20          & 61.5 & 46.8 & 16.1 & 73.1 \\
\bottomrule
\end{tabular}
\end{minipage}
\end{table*}


\section{Memory Scaling}
\label{app:memory_scaling}

\begin{figure}[t]
    \centering
    \includegraphics[width=0.95\columnwidth]
    {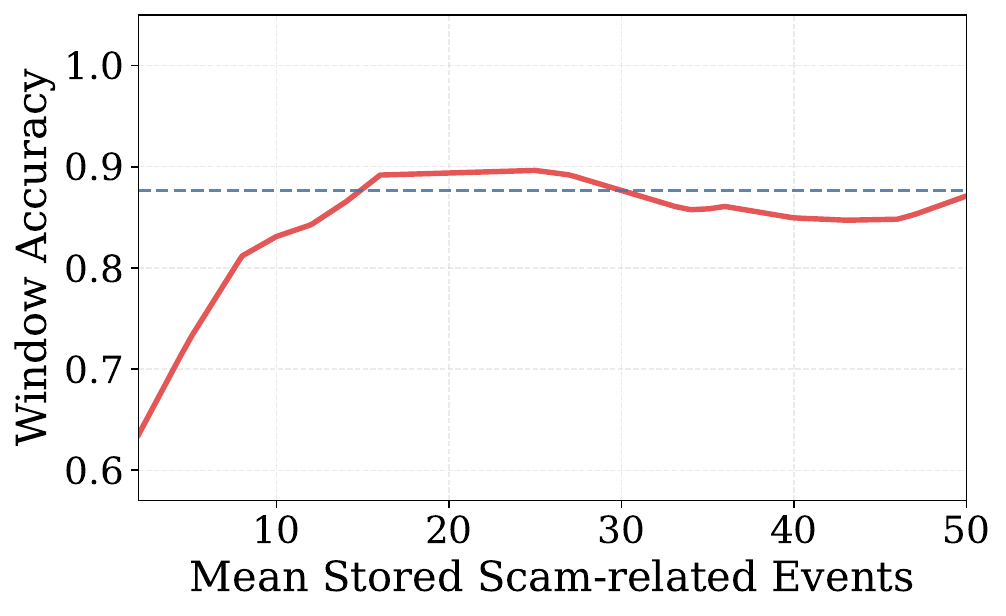}
    \caption{\textbf{Memory scaling analysis.}
    Window-level accuracy as the number of stored scam-related events
    increases.}
    \label{fig:memory}
    \vspace{-2mm}
\end{figure}

Figure~\ref{fig:memory} examines how detection performance changes as
scam-related memory accumulates.
We report window-level accuracy because memory retrieval is performed
independently for each observation window.
Accuracy improves rapidly in the low-memory regime and then gradually
saturates, indicating that historical context is particularly beneficial when
only limited scam evidence has been retained, while additional memory provides
diminishing returns once sufficient relevant evidence has accumulated.

\section{Example of Person-Centric Memory}
\label{app:memory_example}

An example of the structured person record used in our system is shown below.
It demonstrates how person-centric memory stores the entity identifier,
temporal occurrence order, related application history, and associated
metadata.

\begin{tcolorbox}[
enhanced jigsaw,
colback=gray!8,
colframe=gray!50,
boxrule=0.5pt,
arc=2pt,
left=1.5mm,
right=1.5mm,
top=1.5mm,
bottom=1.5mm,
before skip=4pt,
after skip=4pt
]

\textbf{Person ID:} fe28f9515ffd370567f6925b4cb5a17f\\
\textbf{Case ID:} 28886\\
\textbf{Entity Names:} Hayden\\
\textbf{First Seen Order:} 6\\
\textbf{Last Seen Order:} 83\\
\textbf{Related Application History:}

\begin{itemize}
    \item Order 6: \textit{Instant Messaging} was opened.
    \item Order 83: \textit{Instant Messaging} was opened.
    \item ...
\end{itemize}

\end{tcolorbox}


\section{Example of Scam Pattern Reflection Skill}
\label{app:skill_example}

We provide an example of a scam-type-specific skill used for reflection
construction.

\begin{tcolorbox}[
breakable,
enhanced jigsaw,
colback=gray!8,
colframe=gray!50,
boxrule=0.5pt,
arc=2pt,
left=1.5mm,
right=1.5mm,
top=1.5mm,
bottom=1.5mm,
before skip=4pt,
after skip=4pt
]

\textbf{Scam Type:}\\
fake\_online\_investment\_financial\_scam

\vspace{0.4em}
\textbf{Description:}

Scammers lure victims into fraudulent investment schemes through social media
platforms, messaging applications, or malicious financial apps.
These schemes typically promise high returns but ultimately aim to steal the
victim's funds.

\vspace{0.4em}
\textbf{Early Indicators:}

\begin{itemize}
    \item Initial contact initiated via social media platforms with
    investment-related promises.
    \item Scanning QR codes sent by unknown individuals claiming investment
    opportunities.
    \item Downloading suspicious financial or investment applications.
    \item Joining private groups on messaging platforms for so-called
    ``exclusive investment advice.''
    \item Conducting small initial transactions to test withdrawal
    functionality, followed by larger deposits.
\end{itemize}

\vspace{0.4em}
\textbf{Typical App Sequence:}

Social Media
$\rightarrow$
Instant Messaging
$\rightarrow$
Financial/Investment App
$\rightarrow$
Banking App

\vspace{0.4em}
\textbf{Example Reflections:}

\begin{itemize}
    \item In this window, the victim downloaded an unfamiliar investment
    application and initiated contact with a so-called ``customer service''
    representative. The combination of app installation and interaction with
    an unverified individual is a strong indicator of potential fraud.

    \item The victim scanned a QR code from an unknown social-media contact and
    subsequently joined a private messaging group for investment advice.
    These actions suggest early-stage grooming behavior and should raise
    suspicion.
\end{itemize}

\vspace{0.4em}
\textbf{Prompt Enhancement:}

Focus on identifying investment-related app installations, abnormal QR-code
scanning behaviors, and early interactions on social-media or messaging
platforms.
Particular attention should be paid to initial test transactions and
group-joining activities, as these may serve as early warning signals.

\end{tcolorbox}


\section{Sensitivity Analysis}
\label{app:sensitivity}

We examine the sensitivity of ORACLE to two key hyperparameters: the CE-loss
weight $\lambda$ in OPSD and the streaming-window length $W$.
As shown in Table~\ref{tab:sensitivity_ce_window}, removing CE supervision
substantially degrades performance, particularly in EDP and FAR, whereas an
excessively large CE weight also weakens the contribution of privileged
teacher guidance.
A moderate weight of $\lambda=0.1$ achieves the best overall performance.

For window length, very short windows provide insufficient temporal context,
leading to substantially lower HR and PAR.
Performance improves as the observation window becomes longer and peaks at
$W=10$.
Further increasing the window size does not improve performance, suggesting
that excessive local context may introduce redundant or less relevant events.


\section{App Category Taxonomy}
\label{app:taxonomy}

We map exact application names to the following 12 functional categories
throughout benchmark construction:

\begin{tcolorbox}[
breakable,
enhanced jigsaw,
colback=gray!8,
colframe=gray!50,
boxrule=0.5pt,
arc=2pt,
left=1.5mm,
right=1.5mm,
top=1.5mm,
bottom=1.5mm,
before skip=4pt,
after skip=4pt
]
\begin{enumerate}
    \item Communication
    \item Instant Messaging
    \item Social Media
    \item Tools
    \item Financial
    \item Multimedia
    \item Productivity
    \item Travel \& Local
    \item Shopping
    \item Entertainment
    \item Health \& Fitness
    \item Others
\end{enumerate}
\end{tcolorbox}